\documentclass[letterpaper, 10 pt, conference]{ieeeconf}  

\IEEEoverridecommandlockouts  
\usepackage{graphicx} 
\usepackage{lipsum} 
\usepackage{float} 
\usepackage{amsmath} 
\usepackage[biblabel]{cite} 
\usepackage{hyperref}
\hypersetup{%
  breaklinks,
  bookmarksnumbered=true,
  bookmarksopen=true,
  colorlinks=true,
  linkcolor=black,
  urlcolor=black,
  citecolor=black
}

\usepackage{amssymb}
\usepackage[dvipsnames]{xcolor} 
\usepackage{pifont} 

\usepackage{adjustbox}
\usepackage{booktabs}
\usepackage{multirow}
\usepackage{makecell}

\usepackage{cuted} 
\usepackage{caption} 

\usepackage{textcomp}
\usepackage{gensymb} 
\usepackage{svg} 
\usepackage{soul} 
\usepackage{url}
\usepackage{placeins}
\restylefloat{table}

\newcommand{\cmark}{\makecell{\centering {\textcolor{Green}{\ding{52}}}}}
\newcommand{\cmarkstar}{\makecell{\centering {\textcolor{Green}{\ding{52}{\textdagger}}}}}
\newcommand{\xmark}{\makecell{\centering {\textcolor{red}{\ding{55}}}}}

\newcommand{\dataset}{BETTY }
\newcommand{\hours}{13 } 
\newcommand{\size}{32 } 
\newcommand{\spinoutspeed}{63 m/s }

\begin{document}
\bstctlcite{IEEEexample:BSTcontrol}
\title{\dataset Dataset: A Multi-modal Dataset for Full-Stack Autonomy}

\author{Micah Nye$^{1}$, Ayoub Raji$^{2}$, Andrew Saba$^{1}$, Eidan Erlich$^{3}$, Robert Exley$^{4}$,  Aragya Goyal$^{4}$, Alexander Matros$^{3}$,\\Ritesh Misra$^{4}$, Matthew Sivaprakasam$^{1}$, Marko Bertogna$^{2}$, Deva Ramanan$^{1}$, Sebastian Scherer$^{1}$ 
\thanks{$^{1}$ Robotics Institute, Carnegie Mellon University, \texttt{micahn, asaba, msivapra, deva, basti@andrew.cmu.edu}}%
\thanks{$^{2}$ University of Modena and Reggio Emilia, 
\texttt{ayoubraji, marko.bertogna@unimore.it}}
\thanks{$^{3}$ University of Waterloo, \texttt{emerlich, amatros@uwaterloo.ca}}%
\thanks{$^{4}$ University of Pittsburgh, \texttt{robertexley, arg195, rim39@pitt.edu}}%
}


\maketitle
\begin{strip}
  \centering
  \vspace*{-27mm}
  \includegraphics[width=\linewidth]{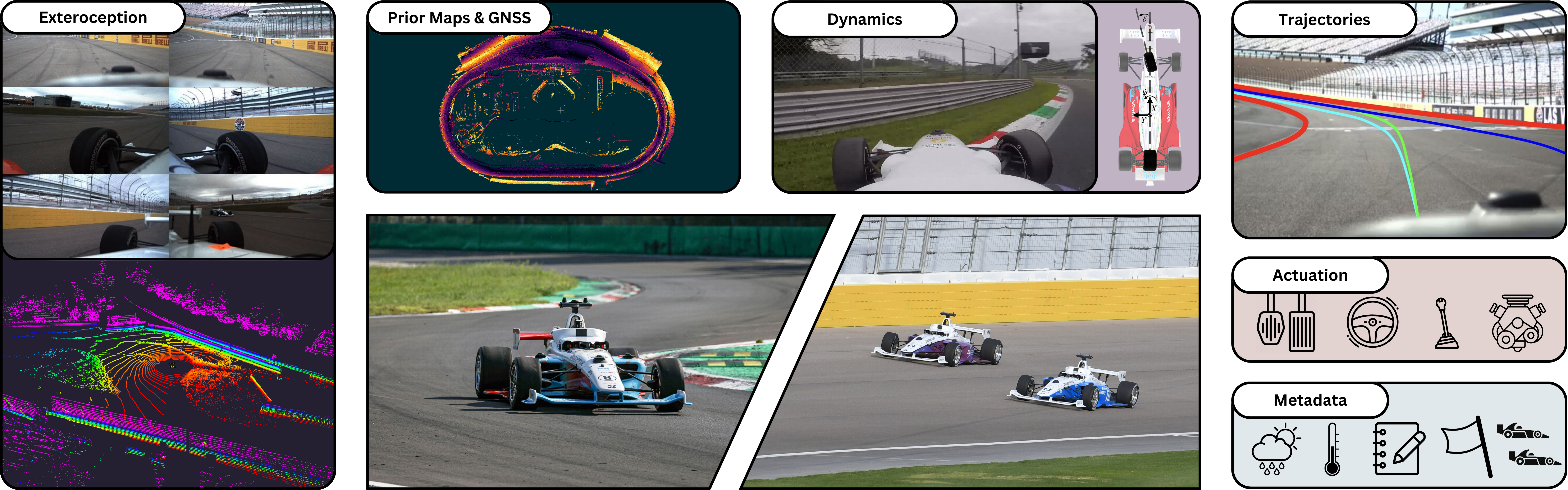}
  \captionof{figure}{We provide exteroceptive sensors (camera, LiDAR, and radar), proprioceptive sensors (tire temperature sensor, slip angle sensor, and more), autonomy data (planned trajectories, actions, and more), and semantic metadata. The \dataset dataset is a tribute to Betty, our autonomous racing vehicle.}
  \label{fig:betty-dataset}
  \vspace*{-1mm}
\end{strip}

\begin{abstract}

We present the \dataset dataset, a large-scale, multi-modal dataset collected on several autonomous racing vehicles, targeting supervised and self-supervised state estimation, dynamics modeling, motion forecasting, perception, and more. Existing large-scale datasets, especially autonomous vehicle datasets, focus primarily on supervised perception, planning, and motion forecasting tasks. Our work enables multi-modal, data-driven methods by including all sensor inputs and the outputs from the software stack, along with semantic metadata and ground truth information. The dataset encompasses 4 years of data, currently comprising over \hours hours and \size TB, collected on autonomous racing vehicle platforms. This data spans 6 diverse racing environments, including high-speed oval courses, for single and multi-agent algorithm evaluation in feature-sparse scenarios, as well as high-speed road courses with high longitudinal and lateral accelerations and tight, GPS-denied environments. It captures highly dynamic states, such as \spinoutspeed crashes, loss of tire traction, and operation at the limit of stability. By offering a large breadth of cross-modal and dynamic data, the \dataset dataset enables the training and testing of full autonomy stack pipelines, pushing the performance of all algorithms to the limits. The current dataset is available at \href{https://pitt-mit-iac.github.io/betty-dataset/.}{https://pitt-mit-iac.github.io/betty-dataset/}.
\end{abstract}


\section{Introduction}

The most challenging robotics tasks involve the integration and interdependence of various modules, such as perception, state estimation, and control. However, without access to a physical robot for development and testing, these tasks cannot be fully developed or validated using available datasets, which typically only offer perception-related data. Unlike the influential KITTI dataset \cite{kitti}, which only has a few sensor modalities, modern datasets include annotations, HD maps, closed-loop planning benchmarks\cite{nuplan, althoff2017commonroad}, ground truth segmentation \& depth \cite{tartanair2020iros}, control signals\cite{sivaprakasam2024tartandrive, kegelman2017revs, Vaseur2019}, and more. Though these datasets enable modern learning-based methods, they alone are still insufficient for addressing the full scope of robotics challenges.

\begin{table*}[ht!]
    \centering
    \renewcommand{\arraystretch}{1.1} 
    \resizebox{\textwidth}{!}{
    \begin{tabular}{@{}cl|lllll|lll|l@{}}
    \toprule
    \multicolumn{1}{l}{\textbf{Data Category}} & \textbf{Modality} & \textbf{KITTI \cite{kitti}} & \textbf{NuScenes \cite{nuscenes2019}} & \textbf{NuPlan \cite{nuplan}} & \textbf{ArgoVerse 2 \cite{argoverse2}} & \textbf{Waymo Open \cite{waymo_open}} & \textbf{TartanDrive 2.0 \cite{sivaprakasam2024tartandrive}} & \textbf{REVS \cite{kegelman2017revs}} & \textbf{RACECAR \cite{racecar}} & \textbf{Ours} \\
    \midrule\midrule
    \multirow{3}{*}{Perception}         & Camera             & \cmark   & \cmark      & \cmark    & \cmark         & \cmark        & \cmark             & \xmark   & \cmark     & \cmark  \\
                                        & LiDAR              & \cmark   & \cmark      & \cmark    & \cmark         & \cmark        & \cmark             & \xmark   & \cmark     & \cmark  \\
                                        & Radar              & \xmark    & \cmark      & \xmark     & \xmark          & \xmark         & \xmark              & \xmark   & \cmark     & \cmark  \\
    \hline
    \multirow{3}{*}{\begin{tabular}{@{}c@{}}State \\ Estimation\end{tabular}}   & IMU                & \cmark   & \cmark      & \cmark    & \xmark          & \xmark         & \cmark             & \cmark  & \cmark     & \cmark  \\
                                        & GNSS               & \cmark   & \cmarkstar & \cmarkstar & \xmark          & \xmark         & \cmarkstar & \cmark  & \cmarkstar & \cmarkstar \\
                                        & Odometry           & \cmark   & \cmark      & \cmark    & \cmark         & \cmark        & \cmark             & \cmark  & \cmark     & \cmark  \\
    \hline
    \multirow{7}{*}{\begin{tabular}{@{}c@{}}Labels \& \\ Metadata\end{tabular}}
                                        & 2D Map             & \cmark   & \cmark      & \cmark    & \xmark          & \cmark        & \xmark              & \xmark   & \xmark      & \cmark  \\ 
                                        & 3D LiDAR Map       & \xmark    & \cmark      & \cmark    & \cmark         & \cmark        & \cmark             & \xmark   & \xmark      & \cmark  \\
                                        & Metadata           & \xmark    & \xmark       & \xmark     & \xmark          & \xmark         & \cmark             & \xmark   & \xmark      & \cmark  \\
                                        & 2D Camera Labels   & \cmark   & \cmark      & \cmark    & \cmark         & \cmark        & \xmark              & \xmark   & \xmark      & \cmark  \\
                                        & 3D Lidar Labels    & \cmark   & \cmark      & \cmark    & \cmark         & \cmark        & \xmark              & \xmark   & \xmark      & \cmark  \\
                                        & Non-ego Odom  & \xmark    & \xmark       & \xmark     & \xmark          & \xmark         & \xmark              & \xmark   & \cmark     & \cmark  \\
                                        & Trajectory & \xmark    & \xmark       & \cmark    & \xmark          & \xmark         & \cmark             & \cmark  & \xmark      & \cmark  \\
    \hline
    \multirow{4}{*}{Actuation}         & Throttle           & \xmark    & \xmark       & \xmark     & \xmark          & \xmark         & \cmark             & \cmark  & \xmark      & \cmark  \\
                                        & Brake              & \xmark    & \xmark       & \xmark     & \xmark          & \xmark         & \xmark              & \cmark  & \xmark      & \cmark  \\
                                        & Steering           & \xmark    & \xmark       & \xmark     & \xmark          & \xmark         & \cmark             & \cmark  & \xmark      & \cmark  \\
                                        & Gear               & \xmark    & \xmark       & \xmark     & \xmark          & \xmark         & \xmark              & \xmark   & \xmark      & \cmark  \\
    \hline
    \multirow{8}{*}{Tire State}         & Engine RPM         & \xmark    & \xmark       & \xmark     & \xmark          & \xmark         & \xmark              & \cmark  & \xmark      & \cmark  \\
                                        & Wheel Torque       & \xmark    & \xmark       & \xmark     & \xmark          & \xmark         & \xmark              & \xmark   & \xmark      & \cmark  \\
                                        & Wheel Speed          & \xmark    & \xmark       & \xmark     & \xmark          & \xmark         & \cmark             & \xmark   & \xmark      & \cmark  \\
                                        & Suspension         & \xmark    & \xmark       & \xmark     & \xmark          & \xmark         & \cmark             & \cmark  & \xmark      & \cmark  \\
                                        & Tire Temp          & \xmark    & \xmark       & \xmark     & \xmark          & \xmark         & \xmark              & \xmark   & \xmark      & \cmark  \\
                                        & Tire Pressure      & \xmark    & \xmark       & \xmark     & \xmark          & \xmark         & \xmark              & \xmark   & \xmark      & \cmark  \\
                                        & Sideslip angle     & \xmark    & \xmark       & \xmark     & \xmark          & \xmark         & \xmark              & \cmark  & \xmark      & \cmark  \\ \bottomrule
    \end{tabular}
    }
    \caption{Comparison of representative population for autonomous driving datasets across urban driving, offroad driving, and racing domains, with full-stack sensors considered. ({\textdagger}) Denotes RTK GNSS.}
    \label{tab:dataset_comparison}
\end{table*}

Furthermore, existing datasets typically target discrete sets of problems. For example, Argoverse 2\cite{argoverse2} targets supervised autonomous vehicle perception, self-supervised LiDAR perception tasks, and motion forecasting. All of these problems focus on building a better understanding of the world, detecting \& tracking the agents in it, and forecasting how the world will evolve. However, this fails to address the challenges of understanding the interaction between the environment and the ego vehicle, particularly in terms of dynamics and state estimation. This issue is even more critical in the car racing domain—both in traditional motorsports and autonomous racing—where vehicles are costly, and data is typically kept confidential, making the field available to limited and privileged research communities. In \cite{remonda2024simulationbenchmarkautonomousracingayoub}, the authors provided a large scale dataset of human- and autonomously-driven racecars in simulation, but is limited to vehicle dynamics and control information, while in the RACECAR dataset\cite{racecar}, only the perception data of an autonomous racecar is provided. The \dataset dataset covers these problems, and more, by including the inputs and outputs of the software stack running on an Autonomous Racing Vehicle (ARV).

By having the sensor inputs and autonomy stack outputs, researchers can evaluate new offline algorithms with a broad range of input sensory data and supervisory signals. To illustrate, \cite{toschi2024lopukf} presents a method that uses LiDAR inertial odometry to estimate the amount of slip experienced by the tires for an ARV, which is used downstream in the dynamics model of a model-based controller. Many evaluations and experiments are not possible with any other dataset due to the lack of cross-functional modalities. For example, with the \dataset dataset, researchers can evaluate how the performance of the localization impacts the slip estimation, which in turn impacts the dynamics model and controller outputs. Additionally, ground truth trajectories, slip sensor readings, and commanded \& actual actuator control signals are included, providing additional metrics to evaluate against.

To summarize, the main contributions of this work are:
\begin{enumerate}
    \item A dataset for developing and evaluating tasks including perception, state estimation, dynamics modeling, motion forecasting, and more using supervised or self-supervised methods.
    \item A dataset with raw sensor data, processed inputs and outputs, metadata, and annotations, that enables evaluating modules independently and chained together in an end-to-end manner.
    \item The most diverse autonomous racing dataset with the largest environment diversity, highest accelerations, and most sensor modalities, by far.
\end{enumerate}


\section{Related Work}

There is a wealth of robotics datasets for many operational domains and environments, including for autonomous vehicles, off-road autonomy, SLAM, vehicle dynamics, and GNSS datasets. The KITTI dataset \cite{kitti} inspired what all modern AV and mobile robotics datasets are today by providing the first large-scale dataset for perception tasks and a suite of evaluation benchmarks.

\subsubsection{Autonomous Vehicle Perception Datasets}
Works such as Argoverse 2, NuScenes, Waymo Open, and ONCE all provide extensive datasets for perception tasks including 3D object detection and tracking and motion forecasting in structured, on-road environments with prescribed traffic rules \cite{argoverse, argoverse2, nuscenes2019, panoptic_nuscenes, waymo_open, mao2021once}. Many more datasets follow suit at a smaller scale to address similar problems or new problems in video object classification, panoptic segmentation, textual semantic understanding, and more \cite{panoptic_nuscenes, RobotCarDatasetIJRR, 2k19dataset, kim2018textualbdd, a2d2}. The RACECAR dataset \cite{racecar}, with open-wheel autonomous racecars, provides high-speed perception \& localization data for 2 oval course environments, but neglects to provide any dynamics information, aside from IMU.

\subsubsection{Autonomous Vehicle Planning Datasets}
Beyond traditional perception tasks, datasets exist for planning and motion forecasting for autonomous vehicles. NuPlan provides a large-scale planning benchmark and an in-house closed- \& open-loop simulator for planning tasks \cite{nuplan}. Waymo \cite{waymo_open}, CommonRoad \cite{althoff2017commonroad}, and MONA \cite{mona2022itsc} also provide high quality object tracks from perception, to be used for planning. 

\subsubsection{Off-road Datasets}
In contrast, there has been a growing presence with off-road datasets in unstructured environments with emphasis on image segmentation, navigation, self-supervised learning, and traversability estimation \cite{sivaprakasam2024tartandrive, rugd, rellis3d, fomo}. 

\subsubsection{SLAM and GNSS Datasets}
Many SLAM datasets exist with the most visually challenging environments with respect to lighting, weather, and geometric \& visual features as well as erratic motion \cite{tartanair2020iros, zhao2024subtmrsdatasetpushingslam, tumrgbd}. While GNSS is also fairly important for localization in the context of many of the papers discussed in this section, very few include any of the raw GNSS observables (e.g. psuedoranges), and instead only provide position and pose estimates from hardware. BETTY is, to the best of our knowledge, the only dataset providing raw GNSS observables at speeds greater than 30 m/s\cite{kaggleGoogleSmartphone}.

\subsubsection{Dynamics Datasets}
Finally, there is a small presence of datasets with extreme dynamics. The Revs Vehicle Dynamics Database \cite{kegelman2017revs} provides a small dataset with dynamics data for vehicle modeling from human expert drivers at speeds in excess of 54 m/s in one static environment. Similarly, the dataset presented in \cite{Vaseur2019} offers data from high-precision sensors capturing force estimation, suspension dynamics, and actuation of a vehicle operating under low-friction conditions.

Table \ref{tab:dataset_comparison} displays related datasets in mobile robotics, with an extensive list of modalities for comparison. The \dataset dataset is the first to combine comprehensive dynamics \& perception data with extensive semantic metadata.


\section{The Dataset}
The \dataset dataset provides as much sensing data and metadata as possible to serve novel multi-modal methods that span beyond traditionally individual modules and datasets. In total, this dataset encompasses perception, vehicle dynamics, motion planning \& forecasting, and SLAM \& GNSS-related tasks for supervised \& self-supervised methods.

\subsection{The Platform}
\begin{figure}
    \centering
    \includegraphics[width=0.7\linewidth]{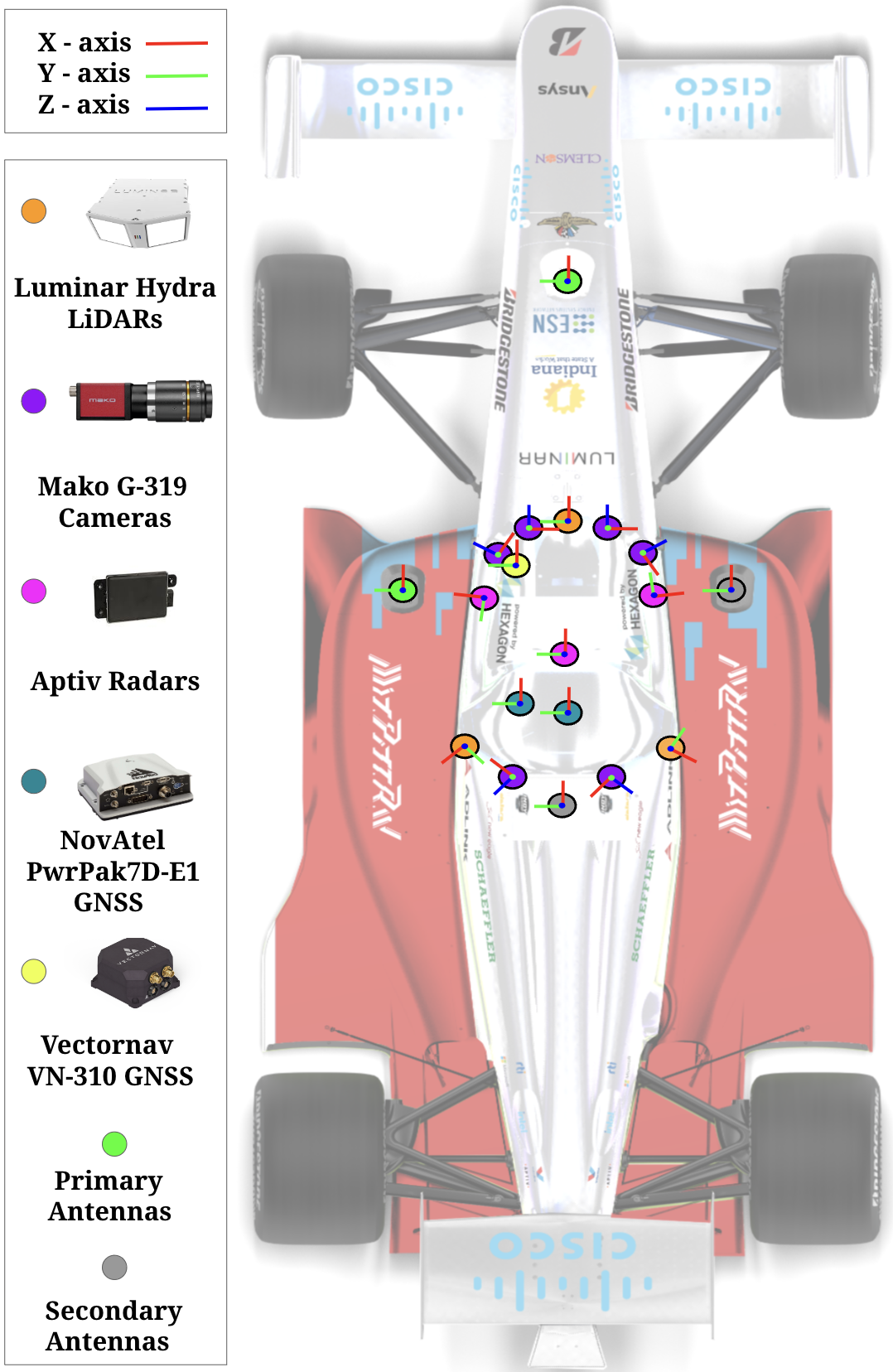}
    \caption{AV-21 sensors and their respective locations.}
    \label{fig:av21_sensors}
\end{figure}

All data is collected from three Dallara AV-21s, the official platform for the Indy Autonomous Challenge (IAC), by three competing teams, MIT-PITT-RW, Unimore Racing, and the Technical University of Munich (TUM), from 2021-2024. The AV-21 is a modified Indy Lights IL-15 chassis \cite{jfr_mprw}, retrofitted with a full autonomy sensor suite, shown in Fig. \ref{fig:av21_sensors}, which includes 3 Luminar Hydra LiDARs, 6 Mako G-319 Cameras, 3 Aptiv Radars, 2 NovAtel PwrPak7D-E1 GNSS units, and, from 2022 onwards, a Vectornav VN-310 GNSS unit. All GNSS units received Real-time Kinematic (RTK) corrections, enabling centimeter-level accuracy. Crucially, a select subset of data also includes a Kistler Correvit SF-Motion non-contact optical sensor for measuring lateral vehicle dynamics, key for determining vehicle stability and sideslip angle. Table \ref{tab:time_sync_sensors} states all sensor rates and synchronization information. In addition, the vehicle has a multitude of vehicle sensors (i.e. tire temperature, etc.) connected over CAN. These sensors combined provide redundant 360$\degree$ coverage with over 200$m$ of sensing range, as well as a complete understanding of tire state, engine performance, drive train, and more. See information in Sec. \ref{sec:calibration} on sensor calibration.

\subsection{Data Collection}
\begin{figure*}[!t]
    \includegraphics[width=\linewidth]{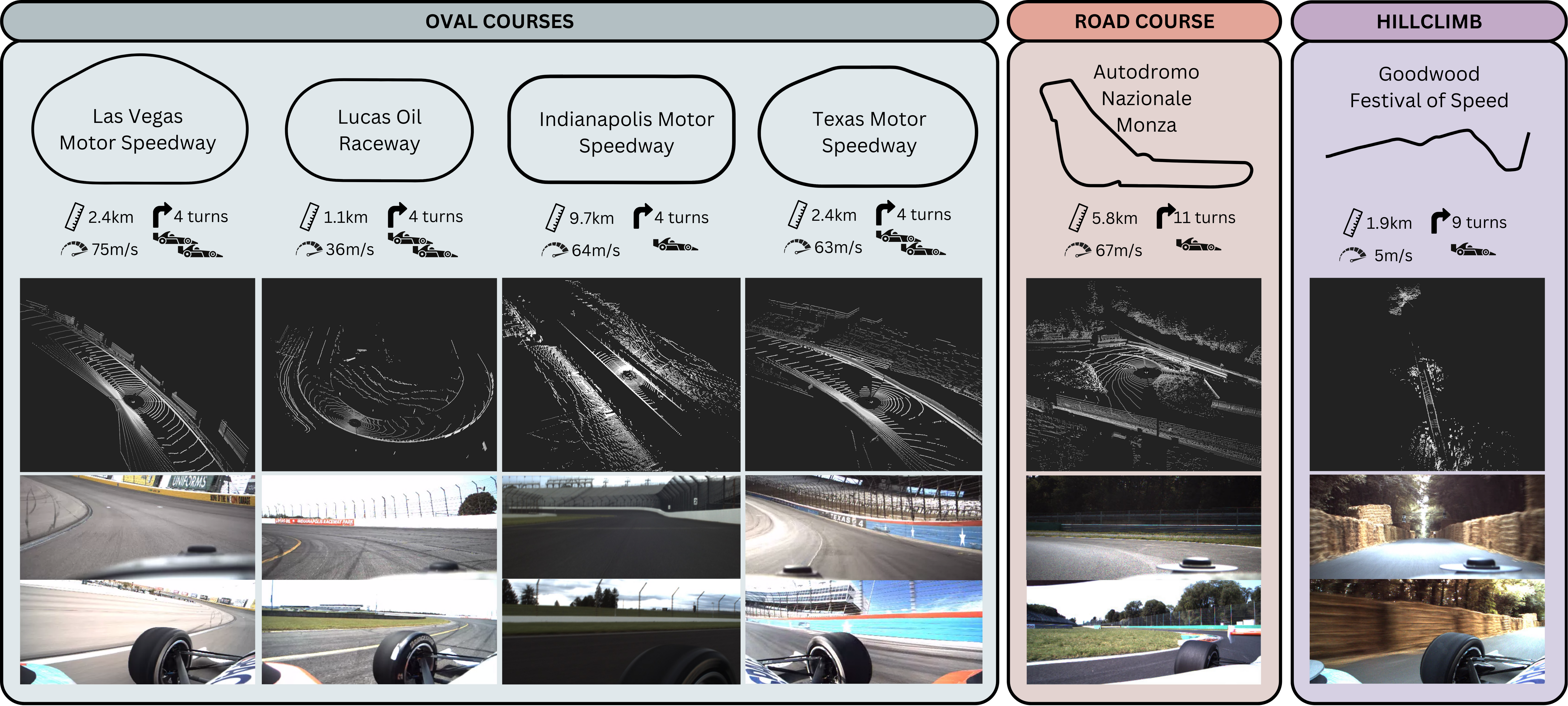}
    \caption{The \dataset dataset was collected in six environments. Each track presents unique characteristics, which present in LiDAR \& camera samples. For example, the oval courses have four large, banked turns. Autodromo Nazionale Monza (Monza) has more turns and high longitudinal accelerations. Goodwood Festival of Speed has tight corridors for navigation.}
    \label{fig:maps_overview}
\end{figure*}

All data was collected during competition or field testing for the IAC. The data was collected across the globe at 4 oval courses, 1 road course, and 1 hillclimb. Fig. \ref{fig:maps_overview} provides an overview of the environments in greater detail.

The data was collected in all seasons, in weather spanning from 3$\degree$C to 32$\degree$C, in cloudy and sunny conditions. Additionally, while no data was collected in the rain due to hardware constraints, data is included from a loss of traction due to a damp track surface.

\subsection{Modalities}
Each run of data provides a wealth of perception, general autonomy \& vehicle state data.

\subsubsection{Perception}
Individual LiDAR pointclouds, camera images, and radar tracks are provided. In addition to this, we provide 360$\degree$ stitched pointclouds as well LiDAR depth-rendered images.

\subsubsection{Autonomy \& State Estimation}
We provide RTK GNSS sensor data, as well as 6-axis and 9-axis IMUs. Commanded and actual throttle, brake, and steering commands are provided, as well as joystick overrides, transmission gear, and powertrain data. Additionally, extensive tire state data, including tire pressure, temperature, speed, torque, and suspension load is provided for tire modeling and vehicle stability. A select amount of runs provide vehicle dynamics sensor data which provide lateral velocity data for side slip angle estimation. 


\subsection{Annotations}
In addition to these modalities, to provide greater utility for users, the dataset was post-processed and annotated with metadata and ground-truth opponent state \& observation labels.

\subsubsection{Metadata}
Static and temporal metadata are provided for self-supervision and ease of use when filtering data to be downloaded. This includes static information such as lighting conditions (e.g. sunny, cloudy, etc.) and ambient temperature, as well as summary information such as maximum \& average speed, single vs multi-agent, opponent, environment information including name \& circuit type, distance traveled, number of passes, if there was a collision, and much more. Testing plan notes are also provided, including the run log from operators, the predefined intent of the run, and the success of the run. More information about the static metadata can be fully described on the website.

Temporal information is included to namely serve as rules-based supervisory signals. This includes race control flags specifying track condition flags which regulate maximum speed, vehicle behaviors, incidents, and if the ego is an attacker or defender.
With this metadata, there is significant external, vehicle-absent data that can be harvested for numerous applications including self-supervision, behavioral learning, and even language processing.

\subsubsection{Labels}

In addition to metadata, we provide post-processed, ground-truth annotated labels of opponent agents. These labels include 2D camera labels and 3D LiDAR labels, and we also provide the ground-truth GNSS position, velocity, and heading for the opponent vehicle. GNSS serves as a robust ground-truth due to RTK, which grants centimeter-level precision.

The camera image labels are generated using Grounding DINO \cite{liu2023grounding}, which takes in as input a camera image and a text prompt (``race car''), and proposes detections in the form of 2D bounding boxes. Post-processing is applied to filter out low-confidence detections and self-detections.

\begin{figure}[t]
    \centering
    \includesvg[width=1.0\linewidth]{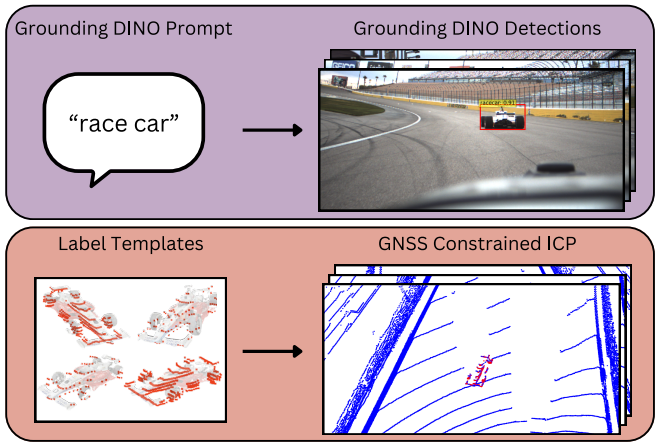}
    \caption{Auto-labelling pipeline. Grounding DINO is used as an initial guess for the camera labeling. For LiDAR corrections, we utilize hand-labels to build a bank of templates that are used to refine an initial guess provided by GPS.}%
    \label{fig:label-gen}
\end{figure}

Automatic 3D LiDAR labels are generated using PointPillars \cite{pointpillars}. If the confidence from PointPillars is below a threshold, we then resort to a template bank from existing hand-labeled point clouds and perform iterative closest point (ICP) alignment with these templates, initialized at a small region around the adversarial GNSS position, as shown in Fig. \ref{fig:label-gen}. A 3D bounding box label is then created using the best matching template. If the error between the GNSS ground-truth and the center of the bounding box is below a specified threshold, the label is accepted as valid. All auto-generated labels are manually verified before being accepted.


\subsection{Dataset Format \& Tooling}
We provide the data for all runs as ROS2 bags as \texttt{mcap} files. ROS2 bags are convenient as ROS2 is an underlying infrastructure for many robotic systems and has a method for real-time playback. The \texttt{mcap} file itself also provides a self-contained way to store all ROS2 bag data, eliminating the need for custom messages to be provided. This improves convenience and open-source compatibility over older bag formats. We also provide a ROS-agnostic format, with \texttt{png}, \texttt{pcd}, and \texttt{csv} files, to facilitate access beyond traditional robotics researchers and reach the classical computer vision, SLAM, and controls \& dynamics modeling communities. All camera and LiDAR annotations are provided in KITTI format \cite{kitti}. The dataset architecture is modular such that researchers can explicitly download desired data. We provide ease-of-use scripts for easy downloading and data merging as well.

\subsection{Calibration}
\label{sec:calibration}
All sensor data was calibrated by Main Street Autonomy (MSA) \cite{msa} with their Calibration as a Service tool. The tool is running visual- \& LiDAR-inertial odometry that considers intra-sweep motion compensation, sensor intrinsics \& extrinsics, and timing errors. This tool calibrates from log data, without fiducials, corner reflectors, or checkerboards. Additionally, using the calibrations, high-definition LiDAR maps can be generated for any run using MSA's Pose as a Service system. See the prior map in Fig. \ref{fig:betty-dataset} as a demonstration of what the system produces.

\subsection{Time Synchronization}


The onboard computer receives GPS NMEA strings via a built-in GPS receiver, which Chrony uses to update the system clock. PTP for Linux synchronizes the system and hardware clocks and broadcasts time onto the network. The LiDAR and camera sensors receive this timing information and update their internal clocks. While the clocks are synchronized, the hardware configuration of the AV21 did not allow for the LiDAR and camera sensors to be triggered in unison. IMU and GNSS sensors are synchronized and timestamped via GPS. Since the system clock is derived from GPS time, synchronization can be verified by comparing GPS timestamps with the system time when data is received. On average, this synchronization difference was 2.8 $ms$, and our system's autonomy modules could operate successfully. Table \ref{tab:time_sync_sensors} has an overview of each sensor with its frame rate, clock, and trigger sources.



\section{Analysis}

The \dataset dataset was collected while fielding several ARVs in an exclusive autonomous racing league, previously accessible only to a select group of researchers. In this section, we analyze this complete, privileged data--now available for all--and discuss numerous works that were developed using similar ARV data, highlighting novel methods, current limitations, and the potential for future advancements.

\begin{table}[h]
   \centering
   \resizebox{\columnwidth}{!}{\begin{tabular}{c|c|c|c}
       Sensor & Frame Rate ($Hz$) & Clock Source & Trigger Source \\
       \midrule \midrule
            Novatel IMU & 125 & GPS & Free Running \\
            Novatel GNSS & 20 & Satellite & GPS \\
            Vectornav IMU & 200 & GPS & Free Running \\
            Vectornav INS & 200 & GPS & Triggered with IMU \\
            Vectornav GNSS & 5 & Satellite & GPS \\
            Camera & 20 & Internal; PTP sync & Free Running \\
            LiDAR & 20 & Internal; PTP sync & Free Running \\ 
            Radar & 20 & System; on arrival & Free Running \\
            Vehicle Sensors & 100 & System; on arrival & Free Running \\
   \end{tabular}}
   \caption{The GNSS units receive time from GPS directly. PTP synchronizes the LiDAR and camera sensors with the onboard system clock. The sensors that stream via CAN are timestamped on arrival. All timestamps are provided with the raw data, allowing the end user to synchronize the sensors. }
   \label{tab:time_sync_sensors}
\end{table}
All multi-agent data was collected on high-speed oval tracks \& was with at most one adversarial agent. Few simultaneous agents and geometrically simple tracks might make the BETTY dataset appear to be less challenging for perception \& motion forecasting tasks at first glance. In reality, however, the high-speed, multi-modal nature of the data promotes the development of methods that are robust under latency constraints at both a software and hardware level. Finally, if new multi-agent data on road courses or data with more agents is made available, the current tooling and dataset format will support it.

\subsection{Perception}
Researchers from TUM presented their perception stack in \cite{tum_preception}, which was developed using similar data. They found that standard, off-the-shelf algorithms \& methods are sufficient, likely due to the structured nature of autonomous racing. Similar to previous findings, \cite{jfr_mprw, Saba-2023-139200, er_autopilot}, the challenge is discovered in system design rather than algorithm development, since standard approaches in object detection \& tracking are effective for autonomous racing. For example, in \cite{tum_latency} researchers analyzed the impact of latency on the ability of the autonomy stack to safely respond to other agents when racing at higher speeds. Similar to \cite{Li2020StreamingP}, there exists a trade-off between the accuracy and latency for autonomous racing perception, which remains one of the largest open challenges. The \dataset dataset enables any researcher to evaluate the end-to-end accuracy and latency of their autonomous racing perception stack. 

Outside of additional work in system design, gaps exist in better utilizing all sensor modalities for object detection \& tracking. Previous works \cite{tum_preception, jfr_mprw, er_autopilot} primarily focus on LiDAR- \& radar-based object detection, with some work on utilizing the cameras. The \dataset dataset, with all sensor modalities \& ground truth information, allows more researchers to explore fusing camera information with LiDARs \& radars, such as \cite{harley2023simple}, \cite{liu2022bevfusion}, \cite{triest2024velociraptor}. New and improved multi-modal methods also provide the system designer with more options. 

\subsection{Motion Forecasting}
Motion forecasting is a critical task in autonomous racing, commonly approached using classical methods like Kalman Filters \cite{er_autopilot}, or more advanced neural networks \cite{tum_prediction}. The key takeaway from prior work is that there is a need to leverage physics when forecasting to constrain the trajectories to a feasible set \cite{tum_prediction}. Additionally, the researchers emphasize the need for fast processing time and efficiency, due to the high-speed and high-acceleration nature of autonomous racing. Despite only targeting high-speed oval racing, with lower max accelerations and simpler track complexity as compared to road courses, the \dataset dataset can enable new research with the interaction of more sensor modalities, semantic metadata, and high-dimensional ground-truth opponent states.


\subsection{Controls \& Vehicle Modeling}

In the literature, control modules for ARVs are typically based on vehicle models with varying levels of complexity. In \cite{tum_control}, a Tube Model Predictive Controller (MPC) is employed, utilizing a simple point-mass model that requires estimation of the vehicle's GG-diagram. In contrast, \cite{kaist_modeling} and \cite{spisak2022robust} implement a Linear Quadratic Regulator (LQR) controller, where a single-track model with a purely linear tire representation is used. More advanced and detailed models are found in \cite{er_autopilot} and \cite{eu_modeling}, where MPC frameworks are applied. The former adopts a dynamic single-track model with the Pacejka Magic Formula tire model, while the latter uses a tricycle model that incorporates the additional yaw moment generated by the rear differential.

\begin{figure}
    \centering
    \includegraphics[width=\linewidth]{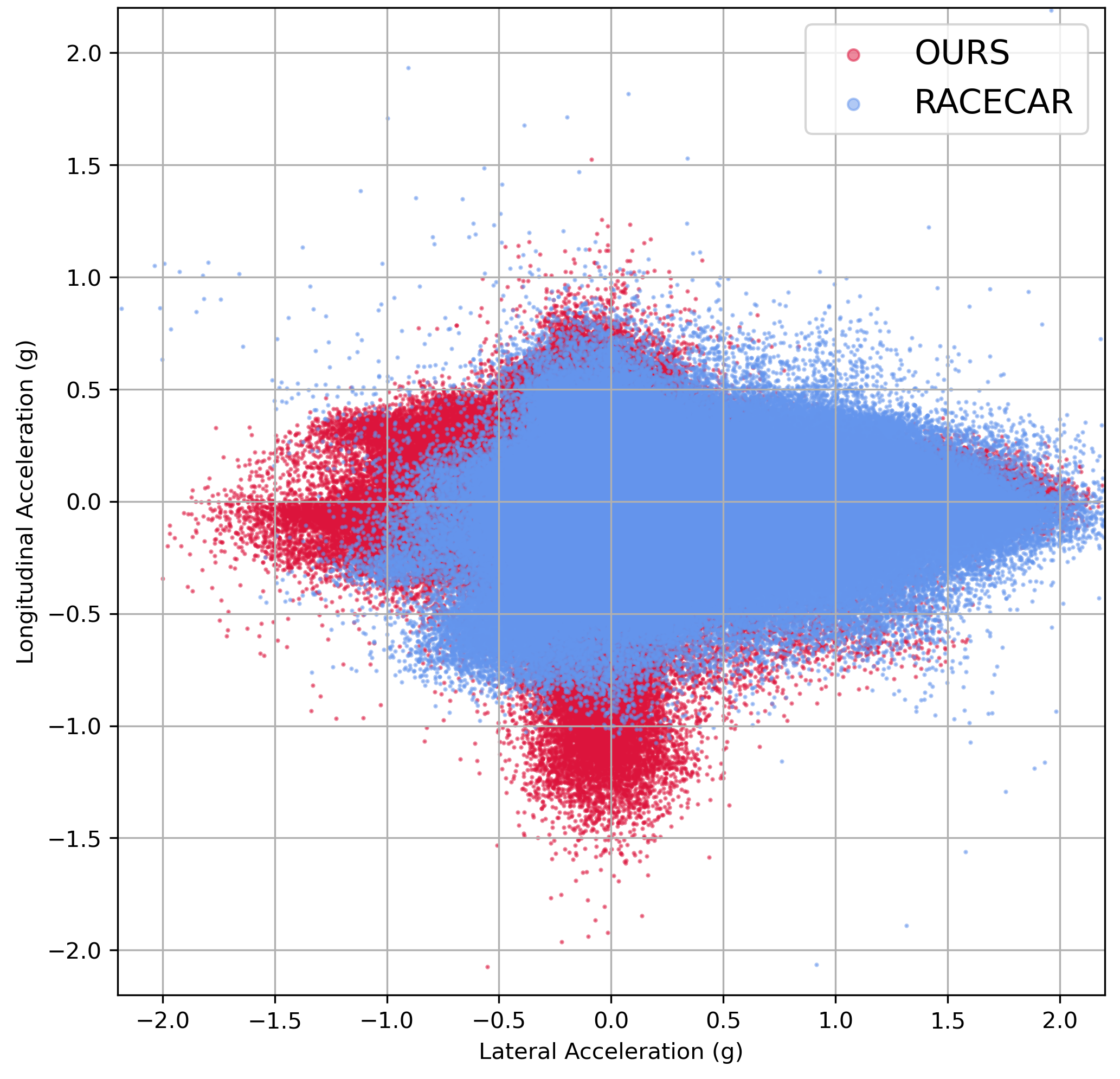}
    \caption{G-G plot of the top 10 most dynamic runs from \dataset and RACECAR. \dataset extends RACECAR's acceleration distribution with additional high braking deceleration and negative lateral acceleration.}
    \label{fig:gg_plot_vs_racecar}
\end{figure}

For these models to be reliable, they must be calibrated using data from the actual vehicle. The \dataset dataset offers data to enable this across diverse scenarios. Notably, it includes performance metrics that exceed those currently reported in the literature, as depicted in Fig. \ref{fig:gg_plot_vs_racecar}. Thanks to the breadth and precision of this dataset, it is now possible to investigate more detailed vehicle models and apply advanced tools for model fitting, leading to enhanced control strategies and state estimation for ARVs.

\subsection{SLAM \& GNSS}

Localization on oval racetracks presents an easier challenge compared to other environments due to the simple nature of the track and the surrounding area, which ensures robust GNSS signals. However, solely relying on GNSS data poses significant risks, as GNSS signals can be susceptible to jamming or temporary hardware failures. To mitigate these risks, alternative localization methods have been explored. For example, in \cite{er_autopilot}, the authors created a LiDAR-based map of the Las Vegas Motor Speedway (LVMS), which they used alongside an Extended Kalman Filter (EKF) to enhance localization robustness. A similar approach was employed for localizing the same ARV at the Monza circuit \cite{toschi2024lopukf}, where traditional GNSS-based localization methods were unreliable due to the presence of bridges and signal occlusions.

To address these challenges without using SLAM or LiDAR Odometry, \cite{tum_localization} proposed a hierarchical state estimation module. This module incorporates an Unscented Kalman Filter (UKF) to estimate the vehicle's sideslip angle as an additional measurement input to the EKF. This approach has been demonstrated to perform effectively on the Monza track, outperforming the Inertial Navigation System (INS) integrated into the GNSS \& IMU receiver.

Other promising solutions that could be evaluated using the \dataset dataset include vision-based odometry algorithms, which have not yet been extensively explored in the autonomous racing domain.

\section{Conclusion}
In this work, we present the \dataset dataset, which was collected on ARVs and significantly extends the capabilities of current open-source, large-scale datasets. Our analysis suggests there are large remaining gaps in achieving robust state estimation, dynamics modeling, motion forecasting, and perception for autonomous racing. Current datasets offer fewer modalities for solving tasks and do not address all of these tasks at once. \dataset is a complete dataset with data to address all these tasks, using supervised or self-supervised methods, with additional annotations, calibrations, and metadata. In the future, we plan to expand the dataset size and work to provide a suite of benchmarks as a uniform evaluation method.



\section{Acknowledgements}


We thank the MIT-PITT-RW team members who have contributed to this work —Arjun Chauhan, Scott Hao, Simerus Mahesh, Jatin Mehta, Andre Slavescu, Raghavesh Viswanath, Shuda Zhong—, for their efforts toward this project.
We would also like to thank Team Unimore Racing —Alessandro Toschi, Nicola Musiu, Micaela Verucchi—for their collaboration on the dataset. We extend a special thanks to Jacob Panikulam, Samarth Wahal, Ethan Eade and Daniel Lu from Main Street Autonomy for providing calibrations and HD maps, enabling us to achieve greater accuracy in our results. Lastly, we would like to extend our thanks to team TUM for contributing data for the Goodwood track. 

\clearpage
{
\bibliographystyle{unsrt}
\bibliography{refs}
}

\end{document}